%% file: paper.tex
\documentclass[english,10pt,letterpaper]{article}
\usepackage{amsmath}
\usepackage{amsfonts}
\usepackage{amssymb}
\usepackage{amsthm}
\usepackage{epsfig}
\usepackage{graphicx}
\usepackage{times}
\usepackage{multirow}
\usepackage{relsize}
\usepackage{url}
\usepackage{cite}
\usepackage{bm}
\usepackage{tikz}
\usetikzlibrary{trees}
\usepackage[algoruled,longend]{algorithm2e}
\usepackage{colortbl}
\usepackage[pagebackref=true,breaklinks=true,letterpaper=true,colorlinks,bookmarks=false]{hyperref}
\newcommand{\myhat}{\bm\hat}
\newcommand{\yc}{\cellcolor[rgb]{1.0,0.4,1.0}}	% lighter
\newcommand{\hl}{\cellcolor[rgb]{0.4,0.8,1.0}} % lighter

\newtheorem{example}{Example}

\newcommand{\placeimage}[2][0.95]{\includegraphics[draft=false,width=#1\textwidth,keepaspectratio]{#2}}
\DeclareMathOperator*{\argmin}{argmin}

\renewcommand{\vec}[1]{\mathbf{#1}}
\begin{document}
\title{
    Illuminant Estimation
	using Ensembles of Multivariate Regression Trees
}
\author{
    Peter van Beek \\
    Cheriton School of Computer Science \\
    University of Waterloo \\
    {\tt\small vanbeek@cs.uwaterloo.ca}
    \and
    R. Wayne Oldford \\
    Department of Statistics \& Actuarial Science \\
    University of Waterloo \\
    {\tt\small rwoldford@uwaterloo.ca}
}
\maketitle

\begin{abstract}
White balancing is a fundamental step in the image processing
pipeline. The process involves estimating the chromaticity
of the illuminant or light source and using the estimate
to correct the image to remove any color cast. Given the
importance of the problem, there has been much previous
work on illuminant estimation. Recently, an approach based
on ensembles of univariate regression trees that are fit
using the squared-error loss function has been proposed and
shown to give excellent performance. In this paper,
we show that a simpler and more accurate ensemble model can be
learned by (i) using multivariate regression trees to take into
account that the chromaticity components of the illuminant
are correlated and constrained, and (ii) fitting each tree
by directly minimizing a loss function of interest---such as
recovery angular error or reproduction angular error---rather
than indirectly using the squared-error loss function as a
surrogate. We show empirically that overall our method leads
to improved performance on diverse image sets.
\end{abstract}

%
% Overall clarity. Is the paper well written and well organized? Does the
% writing enable a substantive evaluation of the work? Is good use
% made of examples to illustrate the problem and solutions?
%
% Provide a clear statement of the problem addressed in the paper.
%
% Provide motivation for the problem.
% Why is the problem interesting? important? challenging?
%
% Review of previous work. Describe the work that other researchers
% have done on this problem and show why these previous approaches
% are inadequate (or put this in a separate section or subsection).
%
% Provide a summary of results.
%
\section{Introduction}

White balancing is a fundamental step in the image processing
pipeline both for digital photography and for many computer
vision tasks. The process involves estimating the chromaticity
of the illuminant or light source and using the estimate to
correct the image to remove any color cast. The goal is to
have the colors in the image, especially the neutral colors,
accurately reflect how these colors appear to the human visual
system when viewing the original scene being imaged.
White balancing is either performed onboard the camera---if
the image delivered by the camera is in JPEG format, for
example---or in a post-processing phase---if the image delivered
by the camera is in the camera's native RAW format.
When done onboard the camera, real-time and space considerations
place additional restrictions on white balancing algorithms.

Given the importance of the problem, there has
been much previous work on illuminant estimation (see
\cite{GijsenijGW2011} for a recent survey). Previous work can
be roughly classified as either static or learning-based.
Static approaches can be applied directly to an image to
estimate the illuminant without the need for a training phase.
Examples of static methods include the gray-world, white-patch,
and shades-of-gray \cite{FinlaysonT2004} algorithms. Static
algorithms are used by the majority of cameras \cite{Deng2011}
as they have the advantage of being fast, albeit at the expense
of accuracy~\cite{ChengPCB2015-CVPR}.

In learning-based approaches, a set of examples is used to
fit---in an offline manner---a predictive model and the model is
used to estimate the illuminant. Examples of learning-based
approaches include (in chronological order):
	learning a probabilistic model based on the distribution of colors
		in an image \cite{FinlaysonHH2001,RosenbergHT2001};
	neural networks \cite{CardeiFB2002};
	support vector regression \cite{XiongF2006};
	learning a probabilistic model of the image formation process
		\cite{GehlerRBMS2008};
	algorithm selection using ensembles of
		classification trees \cite{BiancoCCS2010}
		and maximum likelihood classifiers
		\cite{GijsenijG2011};
	learning a canonical gamut and
		performing gamut mapping \cite{GijsenijGW2010};
	learning a correction to a moment-based algorithm such
		as gray-world \cite{Finlayson2013-ICCV};
	nearest-neighbor regression \cite{JozeD2014};
	$k$-means clustering and voting \cite{BanicL2015}; and
	convolutional neural networks \cite{BiancoCS2015,Shi2016}.
Learning-based algorithms have the advantage of being accurate,
albeit often at the expense of speed~\cite{ChengPCB2015-CVPR}.

Recently, Cheng et al.~\cite{ChengPCB2015-CVPR} proposed
a method based on learning an ensemble of regression
trees---a collection of trees whose individual predictions
are combined to produce a single prediction for a new
example \cite{Breiman1996,Dietterich2000ensemble}. Cheng et
al.'s~\cite{ChengPCB2015-CVPR} method is both fast and accurate,
leading to the best overall performance to date on
diverse image sets. Their method uses (ordinary) univariate
regression trees, where each tree predicts a single response
variable. Thus, to estimate an illuminant $\vec{e} = (r, g,
b)$, Cheng et al.~\cite{ChengPCB2015-CVPR} predict the $r$
chromaticity and the $g$ chromaticity independently and from
these two values the estimate of the $b$ chromaticity can be determined using
$b = 1 - r - g$. As well, they use the (ordinary) squared-error
loss function when fitting each regression tree.

\begin{figure}[thb]
\centering
\placeimage[0.70]{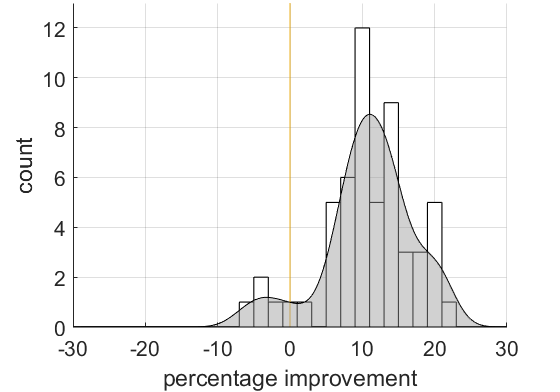}
\caption{
    %Histogram of percentage improvement of our method over Cheng
    Percentage improvement of our method over Cheng
    et al.~\cite{ChengPCB2015-CVPR} for the median of the distance
    measures for an image set, where each image set is captured by
    a single model of a camera, and there are 11 image sets and 5
    distance measures for each image set. For 50 (/55) times our
    method gave an improvement in the median. For 4 (/55) times,
    our method gave an increase in the median; the increases were
    all on one camera, the Canon 600D.
}\label{FIGURE:histogram}
\end{figure}

\subsection{Contributions}

Building on the work of Cheng et al.~\cite{ChengPCB2015-CVPR},
in this paper we make the following contributions.

\begin{enumerate}
\item
We show how multivariate regression trees
\cite{Segal1992,DeAth2002,Larsen2004}, where each tree
predicts multiple responses, can be used to effectively
estimate an illuminant. In the case of multiple responses,
multivariate trees are more compact than univariate trees and
can be more accurate when the response variables are correlated
\cite{Loh2013}. In our method for illuminant estimation,
each tree simultaneously predicts all three chromaticity
components of an illuminant $\vec{e} = (r, g, b)$, rather than
predicting them independently, thus taking into account that
the chromaticity components of the illuminant are correlated
and constrained.

\item
We show how to fit a multivariate regression tree by directly
minimizing a distance measure (loss function) of interest.
Previous work on multivariate regression trees has addressed only
variants of the well-known squared-error loss function
\cite{DeAth2002,Loh2014,Segal1992}. However, for the distance
or performance measures that have been proposed for white balancing,
such as recovery angular error and reproduction
angular error (see, e.g., \cite{FinlaysonZ2014,GijsenijGL2009} and
Section~\ref{SECTION:DistanceMeasures}), the squared error is
not the best surrogate for the distance measure of interest.
We show how to \emph{efficiently} fit multivariate trees for
distance measures that have been proposed for white balancing.

\item
We show empirically that overall our method leads to
improved performance on diverse image sets. Our ensembles
of multivariate regression trees are, with some exceptions,
more accurate (see Figure~\ref{FIGURE:histogram} for a
preliminary presentation of the improvement in accuracy) and
are considerably simpler, while inheriting the fast run-time
of Cheng et al.'s~\cite{ChengPCB2015-CVPR} approach.
\end{enumerate}

\section{Distance Measures}
\label{SECTION:DistanceMeasures}

In this section, we review distance or performance measures that
have been proposed for evaluating the effectiveness of white
balancing algorithms. Although the distance measures are of
course correlated, Finlayson and Zakizadeh \cite{FinlaysonZ2014}
have shown that the ranking of white balancing algorithms will
change depending on the chosen distance measure. The distance
measures assume \emph{normalized} RGB,
\begin{displaymath}
r = \frac{R}{R+G+B}, ~~~
g = \frac{G}{R+G+B}, ~~~
b = \frac{B}{R+G+B}
\end{displaymath}
where $R$, $G$, and $B$ are the red, green, and blue channel
measurements and $r+g+b=1$. In what follows, let
all vectors be row vectors,
$\hat{\vec{e}} = (\hat{r}, \hat{g}, \hat{b})$
be the estimated illuminant, and
$\vec{e} = (r, g, b)$
be the ground truth illuminant.

The most widely used distance measure
\cite{GijsenijGL2009,HordleyF2006},
\emph{recovery angular error},
measures the angular distance between the estimated illuminant and
the ground truth illuminant,
\begin{displaymath}
\mathit{dist}_{\mathit{\theta}}(\myhat{\vec{e}}, \vec{e}) =
	\cos^{-1}
	\left(
	    \frac{\myhat{\vec{e}} \cdot \vec{e}^T}{\|\myhat{\vec{e}}\| \|\vec{e}\|}
	\right) ,
\end{displaymath}
where $\myhat{\vec{e}} \cdot \vec{e}^T$ is the dot product of
the vectors, $\|\cdot\|$ is the Euclidean norm, and we assume
the angular distance is measured in degrees.
Gijsenij, Gevers, and Lucassen \cite{GijsenijGL2009} note
that human observers sometimes judge the recovery angular
error to underestimate the perceived differences between two
images. One reason is that the recovery angular error
ignores the direction of the deviation from the ground truth,
which can be important from a perceptual point of view.

Finlayson and Zakizadeh \cite{FinlaysonZ2014} propose
\emph{reproduction angular error}
as a distance measure, where the angular error
is determined after the image has been corrected with the estimated
illuminant.
Given an estimated illuminant
$\myhat{\vec{e}} = (\myhat{r}, \myhat{g}, \myhat{b})$,
a diagonal correction matrix is given by
$D = \mathrm{diag}(\myhat{g}/\myhat{r},
\myhat{g}/\myhat{g},
\myhat{g}/\myhat{b})$.
%\begin{displaymath}
%D =
%\begin{bmatrix}
%\myhat{g}/\myhat{r} & 0 & 0 \\
%0 & \myhat{g}/\myhat{g} & 0 \\
%0 & 0 & \myhat{g}/\myhat{b}
%\end{bmatrix} .
%\end{displaymath}
Once the correction has been applied to an image,
the angular error between the white balanced ground truth and the
target of uniform gray $\vec{1} = (1, 1, 1)$ can be determined,
\begin{align}\nonumber
\mathit{dist}_{\mathit{rep}}(\myhat{\vec{e}}, \vec{e}) = &
	\cos^{-1}
	\left(
	    \frac{(D \vec{e}^T)^T
		\cdot \vec{1}^T}
		 {\|D \vec{e}^T\|
			\|\vec{1}\|}
	\right) \\\nonumber
      = &
	\cos^{-1}
	\left(
	    \frac{r/\myhat{r} + g/\myhat{g} + b/\myhat{b}}
		{\sqrt{(r/\myhat{r})^2 + (g/\myhat{g})^2 + (b/\myhat{b})^2}
		\; \sqrt{3\vphantom{\myhat{b}^2}}}
	\right) .\nonumber
\end{align}
In other words, given a part of the image that is known to
be achromatic, we are measuring how close in degrees it is to
being achromatic once the image has been white balanced using
the estimated illuminant.

Gijsenij et al.~\cite{GijsenijGL2009} discuss the use of the
Minkowski distance for measuring the distance between the
estimated and the ground truth illuminant. Two special cases
are the \emph{Taxicab distance},
\begin{displaymath}
\mathit{dist}_{1}(\myhat{\vec{e}}, \vec{e}) =
	|\myhat{r} - r| + |\myhat{g} - g| + |\myhat{b} - b| ,
\end{displaymath}
and the \emph{Euclidean distance},
\begin{displaymath}
\mathit{dist}_{2}(\myhat{\vec{e}}, \vec{e}) =
	\sqrt{(\myhat{r} - r)^2 + (\myhat{g} - g)^2 + (\myhat{b} - b)^2} .
\end{displaymath}

Gijsenij et al.~\cite{GijsenijGL2009} note that the Euclidean
distance treats each of the RGB channels uniformly whereas
it is known that the sensitivity of the human eye to perceived
differences varies across color channels.
To this end, Gijsenij et al.~\cite{GijsenijGL2009} define
the \emph{perceptual Euclidean error},
\begin{displaymath}
\mathit{dist}_{\mathit{ped}}(\myhat{\vec{e}}, \vec{e}) =
	\sqrt{ w_{r} (\myhat{r} - r)^2 +
	       w_{g} (\myhat{g} - g)^2 +
	       w_{b} (\myhat{b} - b)^2} ,
\end{displaymath}
where weights $w_{r}$, $w_{g}$, and $w_{b}$ capture this
sensitivity and $w_{r} + w_{g} + w_{b} = 1$. In
experiments, Gijsenij et al.~\cite{GijsenijGL2009}
found that the weight vector $(w_{r}, w_{g}, w_{b}) = (0.21,
0.71, 0.08)$ gave a higher correlation
with the judgment of human observers
than that of
$\mathit{dist}_{\mathit{\theta}}$,
$\mathit{dist}_{1}$, and
$\mathit{dist}_{2}$.

%
% Statement of technical solutions. Provides a clear statement of the
% techniques used to solve the problem. Shows how the problem was
% approached, by what methods and techniques.
%
\section{Our Proposal}
\label{SECTION:OurProposal}

In this section, we present our approach for white balancing
based on ensembles of multivariate regression trees. We
begin with a brief review of univariate (ordinary)
regression trees and how they were applied by Cheng et
al.~\cite{ChengPCB2015-CVPR} for white balancing.

\subsection{Univariate regression trees}
\label{SUBSECTION:Univariate}

Univariate regression trees are constructed in
a greedy, top-down manner from a set of labeled training examples
$\{(\vec{x}_1, y_1), \ldots, (\vec{x}_n, y_n)\}$, where
$\vec{x}_i = (x_{i1}, \ldots, x_{im})$ is a vector of feature
values and $y_i$ is a scalar response variable
(see, e.g., \cite{BreimanFSO1984,Quinlan1986,HastieTF2009}).
As it is sufficient for our purposes, we assume that the feature
values and the response variable are real-valued.
The root node of the tree is associated with all the training examples.
At each step in the construction of the tree,
the training examples at a node are partitioned by
choosing the feature $j \in \{1, \ldots, m\}$ and
partition $p$ that minimizes the total of the
squared-error loss functions,
\begin{equation}
\label{EQUATION:meanSquareErrorTrees}
    \argmin_{c_1} \mathlarger{\sum}_{i \in L(j,p)}
	( y_i - c_1 )^2
    ~+~
    \argmin_{c_2} \mathlarger{\sum}_{i \in R(j,p)}
	( y_i - c_2 )^2 ,
\end{equation}
where the partition is into two subsets, $L(j,p)$ and $R(j,p)$,
formed by branching on feature $j$ according to $x_{ij} \leq
p$ and $x_{ij} > p$, respectively. For the
squared-error loss function, for any choice of feature $j$
and partition $p$ the minimization is solved by taking
the scalar $c_1$ to be the mean of the $y_i$ values in the
left branch and the scalar $c_2$ to be the mean of the $y_i$
values in the right branch (see, e.g., \cite{HastieTF2009}).
Once the best pair ($j, p)$ is found, a left child node and
right child node are added to the tree and are associated with
the subsets $L(j,p)$ and $R(j,p)$. The partitioning continues
until some stopping criterion is met, in which case the node
is a leaf and is labeled with the scalar~$c$ associated with the
subset of examples at the node.

To estimate an illuminant $\vec{e} = (r, g, b)$, Cheng et
al.~\cite{ChengPCB2015-CVPR} predict the $r$ chromaticity and
the $g$ chromaticity independently using separate univariate trees
fit with the squared-error loss function
and from these two values
the estimate of the $b$ chromaticity can be determined using $b = 1 - r - g$.
Cheng et al.~\cite{ChengPCB2015-CVPR} use four pairs of simple features
in their univariate trees:
($f_{r}^{1}$,~$f_{g}^{1}$),
the mean color chromaticity as
provided by the gray-world algorithm;
($f_{r}^{2}$,~$f_{g}^{2}$),
the brightest color chromaticity,
an adaptation of the
white-patch algorithm;
($f_{r}^{3}$,~$f_{g}^{3}$),
the bin average of the mode of the RGB histogram, and
($f_{r}^{4}$,~$f_{g}^{4}$),
the mode of the kernel density estimate from the
normalized chromaticity plane.
An important contribution of their work is a novel method for
training and combining the predictions of an ensemble of trees
using these features that is both fast and accurate, leading
to the best overall performance to date on diverse image sets.

Two important points are that Cheng et
al.'s~\cite{ChengPCB2015-CVPR} method
(i) predicts the $r$ and $g$ chromaticities \emph{independently}, and
(ii) minimizes the distance measure of interest---in their work,
the recovery angular error---\emph{indirectly} by minimizing
the squared-error loss function. Our starting point is (i)
the observation that our response variable is not a scalar
but a vector $\vec{e} = (r, g, b)$ of chromaticities, and (ii)
the related observation that independently fitting using the
squared-error loss function is not necessarily a good surrogate
for minimizing distance measures used in white balancing
(see Example~\ref{EXAMPLE:surrogate}).

\begin{example}
\label{EXAMPLE:surrogate}
Let $\vec{e} = (r, g, b)$ be the ground truth illuminant.
Consider the two estimates of the illuminant
$\myhat{\vec{e}}_1 = (\myhat{r}_{1}, \myhat{g}_{1}, \myhat{b}_{1})$ and
$\myhat{\vec{e}}_2 = (\myhat{r}_{2}, \myhat{g}_{2}, \myhat{b}_{2})$, where
\begin{align*}
\myhat{r}_{1} &= r + \alpha,   & \myhat{r}_{2} &= r + \alpha, \\
\myhat{g}_{1} &= g + \alpha,   & \myhat{g}_{2} &= g - \alpha, \\
\myhat{b}_{1} &= 1 - \myhat{r}_{1} - \myhat{g}_{1} = b - 2 \alpha, &
\myhat{b}_{2} &= 1 - \myhat{r}_{2} - \myhat{g}_{2} = b,
\end{align*}
and $\alpha > 0$ is some residual error.
Here, the pair of estimates $\myhat{r}_{1}$ and $\myhat{r}_{2}$
and the pair of estimates
$\myhat{g}_{1}$ and $\myhat{g}_{2}$ both have
equal squared error ($\alpha^2$). Thus, from the point of view of
independently fitting
using the
squared-error loss function, the two estimates of
the illuminant have equal error.
However, for distance measures that have been proposed for
white balancing (see Section~\ref{SECTION:DistanceMeasures}),
the error of the estimate
$\myhat{\vec{e}}_1$ is larger (and can be much larger) than that of
$\myhat{\vec{e}}_2$.
For example, let $\vec{e} = (\frac{1}{3}, \frac{1}{3}, \frac{1}{3})$
and $\alpha = 0.05$.
For the recovery angular error,
$\mathit{dist}_{\mathit{\theta}}(\myhat{\vec{e}}_1, \vec{e}) = 11.977$
and
$\mathit{dist}_{\mathit{\theta}}(\myhat{\vec{e}}_2, \vec{e}) = 6.983$.
Thus, minimizing the squared error is not necessarily a good surrogate
for minimizing a distance measure of interest in white balancing.
\end{example}

\begin{figure}[th!]
\input{exampleTree.tex}
%\placeimage[0.45]{Figures/Tree/tree.png}
\centering
\begin{tabular}{@{}cccc@{}}
%\placeimage[0.22]{Figures/Tree/Leaf-Top/Canon1DsMkIII_0038.jpg} &
%    \placeimage[0.22]{Figures/Tree/Leaf-Top/Canon1DsMkIII_0083.jpg} &
%    \placeimage[0.22]{Figures/Tree/Leaf-Top/Canon1DsMkIII_0154.jpg} &
%    \placeimage[0.22]{Figures/Tree/Leaf-Top/Canon1DsMkIII_0194.jpg} \\
\placeimage[0.22]{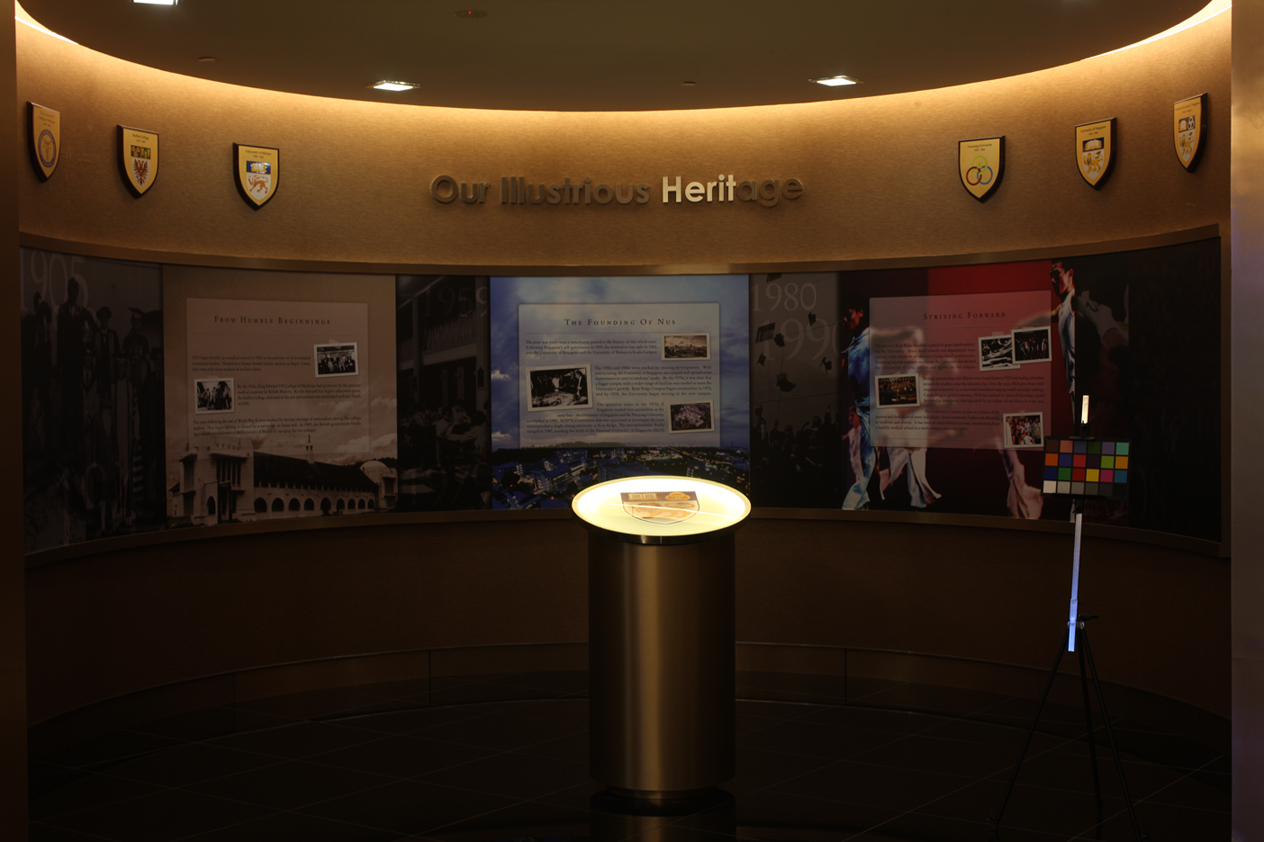} &
    \placeimage[0.22]{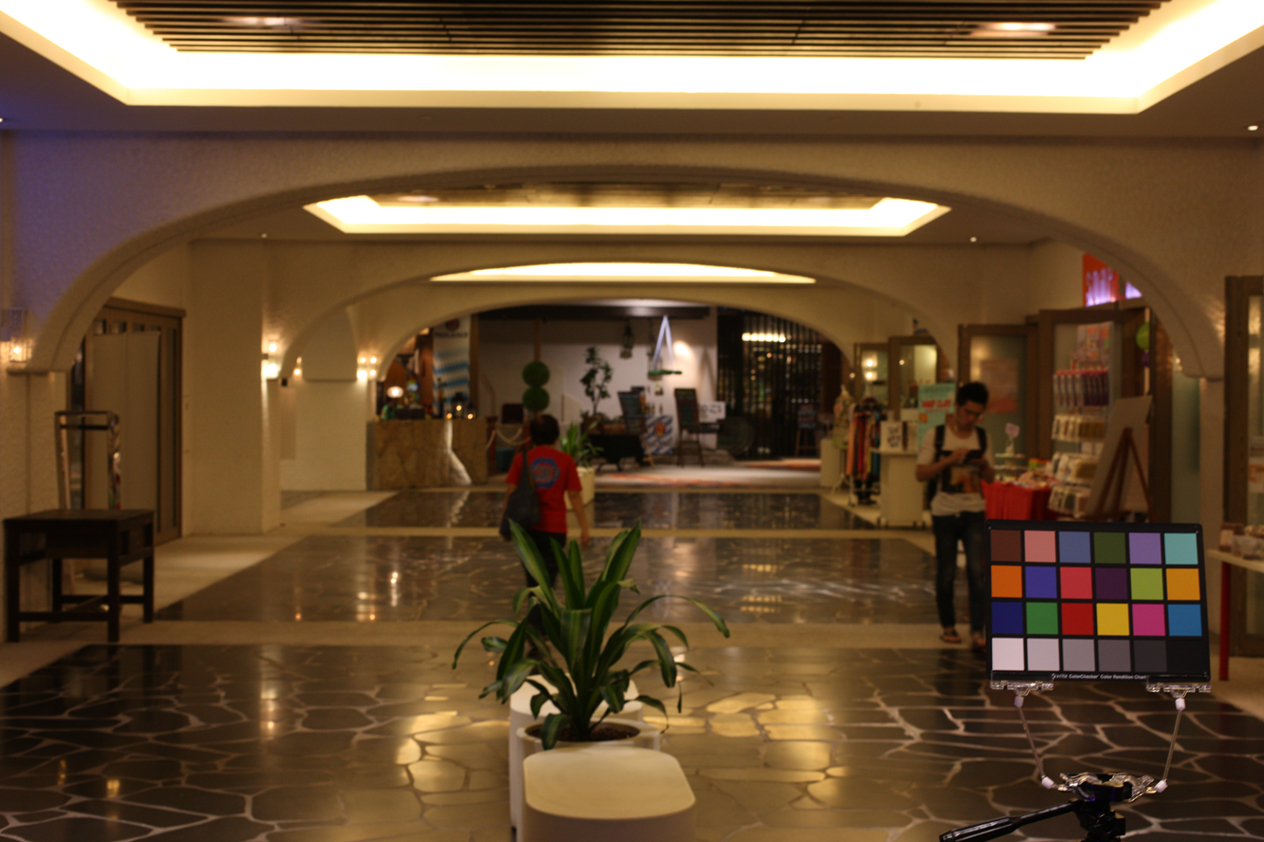} &
    \placeimage[0.22]{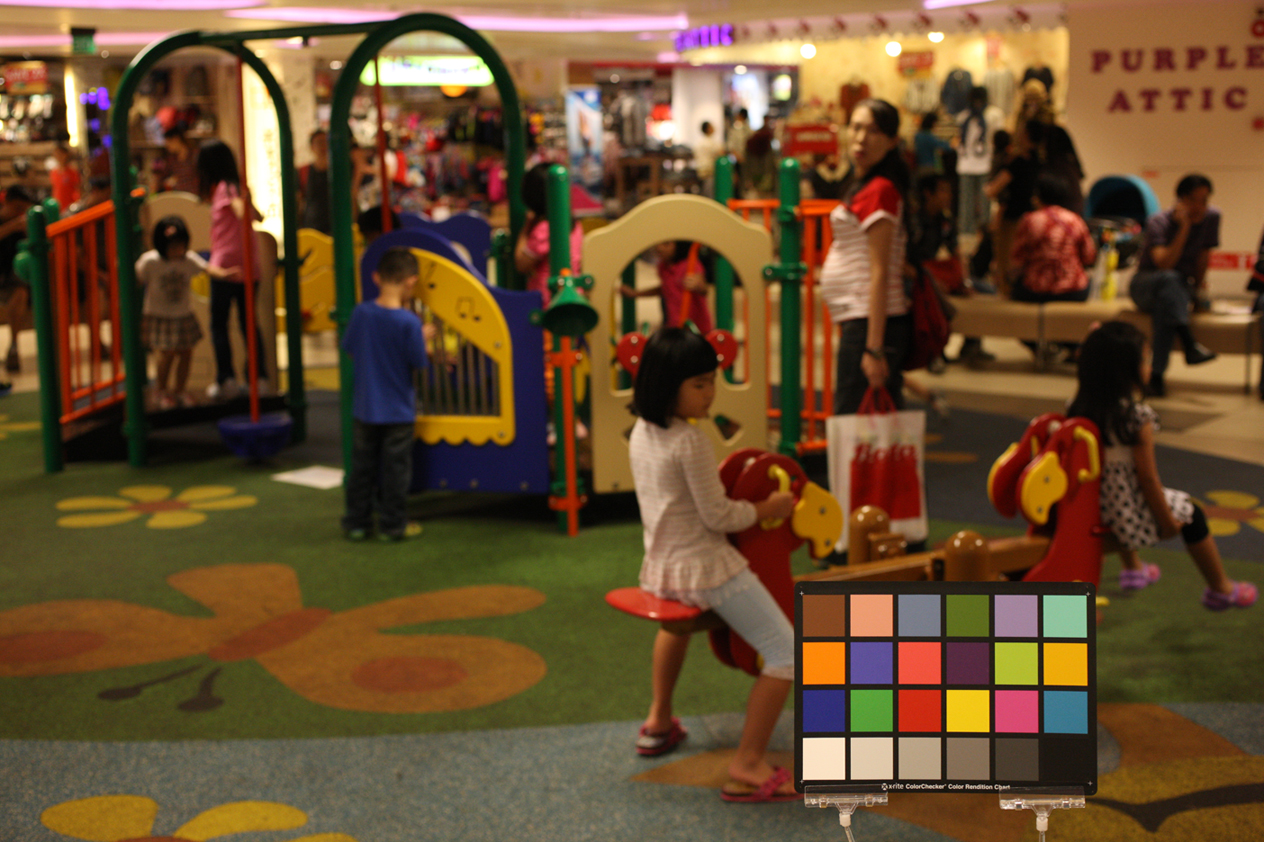} &
    \placeimage[0.22]{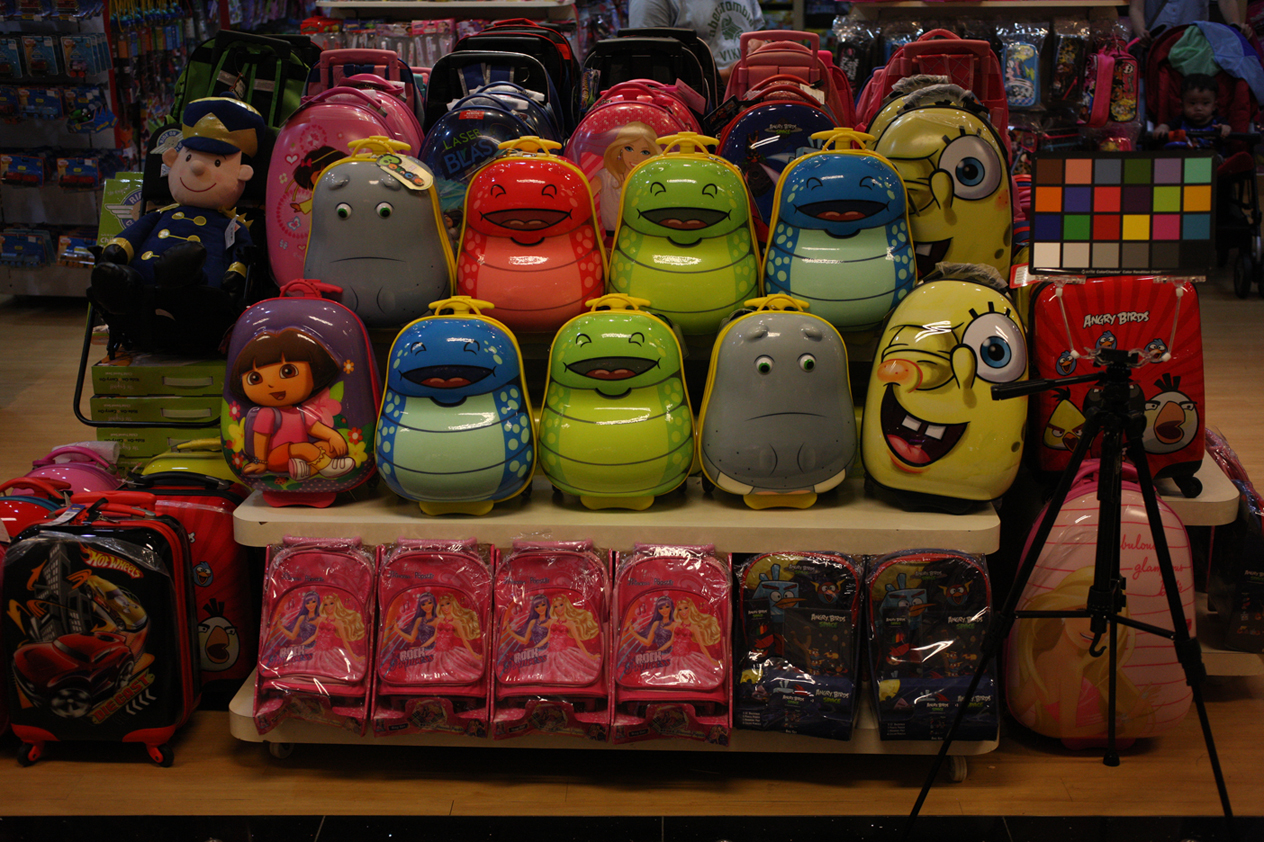}
\end{tabular}
\caption{
    An example multivariate regression tree when the distance measure is
    recovery angular error, showing only the first four layers
    (out of 18 layers). The three (green) leaf nodes are labeled
    with the number of training examples at that node and the
    illuminant estimation if that node is reached. Also shown are
    the four images associated with the bottom-most leaf node;
    the images have been gamma corrected for presentation.
}\label{FIGURE:tree}
\end{figure}

\subsection{Multivariate regression trees}

Multivariate regression trees are constructed in a greedy,
top-down manner from a set of labeled training examples
$\{(\vec{x}_1, \vec{y}_1), \ldots, (\vec{x}_n, \vec{y}_n)\}$,
where as before $\vec{x}_i$ is a vector of feature values
but now $\vec{y}_i = (y_{i1}, \ldots, y_{ik})$ is a \emph{vector} of
response variables \cite{Segal1992,DeAth2002,Larsen2004}.

Our proposed method for illuminant estimation is for a
single tree to simultaneously predict all three chromaticity
components of an illuminant $\vec{e} = (r, g, b)$, rather than
multiple trees predicting them independently, thus taking into
account that the chromaticity components of the illuminant are
correlated and constrained (see Figure~\ref{FIGURE:tree}).
An innovation of our approach is to fit a multivariate
regression tree by directly minimizing a distance measure
(loss function) of interest. Previous work on multivariate
regression trees has addressed only loss functions that
are variants of the well-known squared-error loss function
\cite{DeAth2002,Segal1992}. In our proposed method, at each
step in the construction of the multivariate regression tree,
the training examples at a node are partitioned by choosing
a feature $j \in \{1, \ldots, m\}$ and partition $p$ that
minimizes the total of the loss functions,
\begin{equation}
\label{EQUATION:distErrorTrees}
    \argmin_{\myhat{\vec{e}}_1}
	\mathlarger{\sum}_{i \in L(j,p)}
	    \mathit{dist}(\myhat{\vec{e}}_{1}, \vec{e}_{i})
    ~+~
    \argmin_{\myhat{\vec{e}}_2}
	\mathlarger{\sum}_{i \in R(j,p)}
	    \mathit{dist}(\myhat{\vec{e}}_{2}, \vec{e}_{i}) ,
\end{equation}
where the partition is into two subsets, $L(j,p)$ and $R(j,p)$,
formed by branching on feature $j$ according to $x_{ij} \leq p$
and $x_{ij} > p$, respectively; $\mathit{dist}(\cdot,\cdot)$ is
a distance measure that has been proposed for
white balancing (see Section~\ref{SECTION:DistanceMeasures});
and $\myhat{\vec{e}}_{1}$, $\myhat{\vec{e}}_{2}$, and $\vec{e}_{i}$
are normalized RGB vectors in $\mathbb{R}^3$, i.e., the
chromaticities sum to one. The partitioning continues until some
stopping criteria is met, in which case the node is a leaf and
is labeled with the estimated illuminant $\myhat{\vec{e}}$
associated with the subset of examples at the node. When such
a tree is used in white balancing an image that has not been
seen before, one starts at the root and repeatedly tests the
feature at a node and follows the appropriate branch until
a leaf is reached. The label of the leaf is the estimated
illuminant of the image.

\subsection{Ensembles of multivariate regression trees}
\label{SUBSECTION:Ensembles}

To improve predictive accuracy, an ensemble of trees
is learned where each tree is a multivariate regression
tree. Various methods have been proposed for constructing
ensembles including manipulating the training examples,
manipulating the input features, injecting randomness into
the learning algorithm, and combinations of these methods
(see \cite{Dietterich2000ensemble}). We adopt the method
for constructing ensembles that injects randomness into the
construction of an individual tree.
When constructing the tree,
each feature $j$ partition $p$ pair considered in
Equation~\ref{EQUATION:distErrorTrees}
is accumulated
along with the total of the loss functions
associated with the pair. Rather than taking the $(j, p)$
pair that minimizes the total of the loss functions, a $(j,
p)$ pair is chosen at random from all of the pairs that are
within a given percentage of the minimum. This ensures
that diverse yet accurate trees are constructed. We also
experimented with bagging \cite{Breiman1996} and Cheng et al.'s
method~\cite{ChengPCB2015-CVPR}, two methods that manipulate
the training examples to construct diverse trees, and
with random forests \cite{Ho1995,Breiman2001}, a method
that manipulates both the training examples and the input
features. However, in each case the alternative led to a
decrease in accuracy.

\begin{figure*}[t]
\centering
\placeimage[1.00]{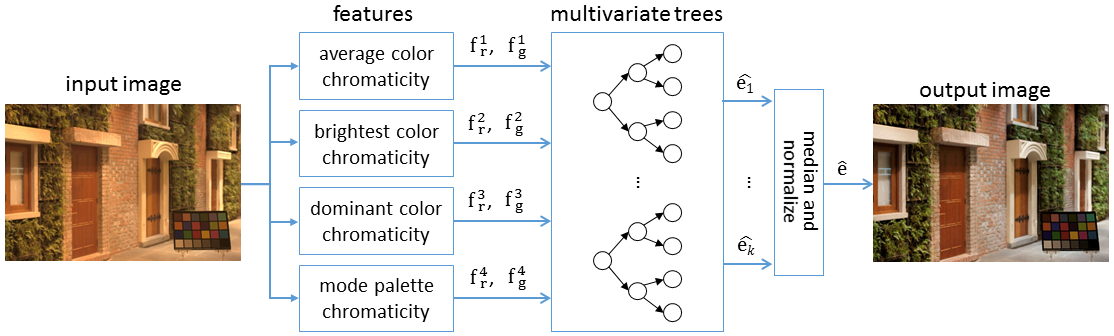}
\caption{
    Our adaptation of Cheng et al.'s~\cite{ChengPCB2015-CVPR}
    learning-based method. Given an
    input image, four pairs of feature values are calculated, the
    feature values are used by each tree in the ensemble to predict
    the illumination, and the individual predictions
    $\{\myhat{\vec{e}}_{1}, \ldots, \myhat{\vec{e}}_{k}\}$
    are combined
    into a single estimate
    $\myhat{\vec{e}}$
    of the illuminant that is used to white balance the image.
}\label{FIGURE:pipeline}
\end{figure*}

Given a new image, the individual predictions of the trees in
the ensemble are combined into a single estimated illuminant
$\myhat{\vec{e}}$ for the image as follows
(see Figure~\ref{FIGURE:pipeline}).
Let $S = \{\myhat{\vec{e}}_{1}, \ldots, \myhat{\vec{e}}_{k}\}$
be the set of predictions from
the individual trees. The final estimated illuminant
is the normalized RGB vector
$\myhat{\vec{e}} = (\myhat{r}, \myhat{g}, \myhat{b})$
that minimizes
$\sum_{i \in S} \mathit{dist}(\myhat{\vec{e}}, \myhat{\vec{e}}_{i})$;
i.e., the same minimization problem as the one that
is solved in partitioning the training examples at a
node and in labeling a leaf when constructing the trees
(Equation~\ref{EQUATION:distErrorTrees}).

\subsection{Fitting multivariate regression trees}
\label{SUBSECTION:fitting}

In fitting a multivariate regression tree to a set of
labeled training examples, the idea is to repeatedly, in
a greedy and top-down manner, find the best feature $j$
and partition $p$ that results in the largest drop in the
total error, as measured by a given distance measure (see
Equation~\ref{EQUATION:distErrorTrees}).
The minimization in Equation~\ref{EQUATION:distErrorTrees}
is a constrained, non-linear optimization problem. The problem
must be solved many times when learning the ensemble of trees
(approximately 40,000 calls to the minimization routine
is typical for constructing a single tree)
and also once every time the ensemble is applied to a new image and the
individual predictions of the trees are combined to obtain a
final estimated illuminant. Note that learning the trees is an
offline process that occurs once while applying the ensemble
to a new image is an online process that will occur many times,
as it occurs each time an image is captured by the camera.

The minimization problem can be solved using \emph{exact}
but sophisticated and computationally expensive numerical
optimization routines such as those based on interior-point
or sequential quadratic programming methods. Exact methods
are suitable for offline learning of a single ensemble of
multivariate trees, but unfortunately are too slow for
experimental evaluation where cross-validation and
repeated trials are necessary to obtain accurate estimates of
performance. This impediment also arises in univariate trees
when using loss functions other than squared error (see, e.g.,
\cite[p.~342]{HastieTF2009}). As well, sophisticated exact
methods are unlikely to be suitable for online application
of the ensemble onboard the camera, given the real-time and
space considerations.

Our proposed method is to solve the minimization problem
\emph{approximately} by taking the median of the ground
truth illuminants of the training examples at a node
and normalizing so that the RGB values sum to one
(see Example~\ref{EXAMPLE:approximate}). The method
is simple and fast.
As we show in supplementary material,
the method often finds the exact solution and otherwise
finds good quality approximate solutions
for distance measures that arise in white balancing.
Further, solving the minimization problem approximately
does not appear to have a significant impact on the accuracy
of the multivariate regression trees, perhaps because the
tree construction itself is a greedy (and hence approximate)
process.

%
% norm2 error
%
% rgb    (0.4284847, 0.4468358, 0.1246795)
% rgb    (0.4220675, 0.4472919, 0.1306407)
% rgb    (0.4098007, 0.4682081, 0.1219912)
% yMean   (0.4201176, 0.4541119, 0.1257705)
% yMedian (0.4220675, 0.4472919, 0.1246795), denom = 0.9940389
% c       (0.4245986, 0.4499742, 0.1254272)
% cExact  (0.4227879, 0.4487327, 0.1284794)
% approx = 3.5169881, exact = 3.4092286, mean cost = 3.7618532, per = 3.1608184
% 
% 4 digits accuracy.
% 
% e1:      0.4285,  0.4468,  0.1247
% e2:      0.4221,  0.4473,  0.1306
% e3:      0.4098,  0.4682,  0.1220
% median:  0.4221,  0.4473,  0.1247
% approx:  0.4246,  0.4500,  0.1254
% exact:   0.4228,  0.4487,  0.1285
% approximate cost =  3.5134
% exact cost       =  3.4049
% 
% 5 digits accuracy.
% 
% e1:     0.42848, 0.44684, 0.12468
% e2:     0.42207, 0.44729, 0.13064
% e3:     0.40980, 0.46821, 0.12199
% median: 0.42207, 0.44729, 0.12468
% approx: 0.42460, 0.44997, 0.12543
% exact:  0.42279, 0.44873, 0.12848
% approximate cost = 3.51646
% exact cost       = 3.40892
%
\begin{example}
\label{EXAMPLE:approximate}
Consider the follow three training examples, where the ground
truth illuminants are shown but feature values are not
shown, and let $\mathit{dist}_{2}$ be the distance measure.
\begin{center}
\begin{tabular}{c|lll}
example &
    \multicolumn{1}{|c}{r} &
    \multicolumn{1}{c}{g} &
    \multicolumn{1}{c}{b} \\\hline
$\vec{e}_{1}$ & 0.4285 & 0.4468 & 0.1247 \\
$\vec{e}_{2}$ & 0.4221 & 0.4473 & 0.1306 \\
$\vec{e}_{3}$ & 0.4098 & 0.4682 & 0.1220
\end{tabular}
\end{center}
Taking the median of the training examples gives
(0.4221, 0.4473, 0.1247) and normalizing so that the
RGB values sum to one results in the illuminant estimate
(0.4246, 0.4500, 0.1254) with an associated cost of 3.5134.
Taking the mean of the training examples
results in the illuminant estimate
(0.42012, 0.45411, 0.12577) with an associated cost of 3.7619.
The illuminant estimate that exactly minimizes the sum of the
distance measure over the training examples is
(0.4228, 0.4487, 0.1285) with an associated cost of 3.4049.
\end{example}

%
% Evaluation of the proposal. Shows how successful our solutions were,
% and why they were successful. Provide convincing evidence in support
% of all claims. Carefully evaluate the strengths and limitations
% of the contribution. Some dimensions for evaluation include empirical
% results and theoretical analyses.
%
\section{Experimental Evaluation}
\label{SECTION:ExperimentalEvaluation}

In this section, we compare a MATLAB implementation
of our multivariate regression tree method to Cheng et
al.'s~\cite{ChengPCB2015-CVPR} method\footnote{The software is available at:
\url{https://cs.uwaterloo.ca/~vanbeek/Research/research_cp}
}.

\subsection{Image sets}

We used the following image sets in our experiments.

\emph{SFU Laboratory image set.}
The SFU Laboratory image set consists of images of objects in
a laboratory setting captured by a video camera
under 11 different illuminants \cite{BarnardMCF2002}.
Following previous work, we use a
common subset of 321 images (the minimal specularities and
non-negligible dielectric specularities subset).
A single image is the average of 50 video frames
and is stored as a 16-bit TIFF file. In contrast to more recent
image sets, the images are not captured as a camera RAW image file
and thus have been processed by the camera.
A white reference standard was used to determine ground truth,
but the white reference does not appear in the images.

\emph{Gehler-Shi image set.}
The Gehler-Shi image set consists of 568 indoor and
outdoor images captured by two different cameras
\cite{GehlerRBMS2008,ShiF}. As done in previous work, we used
the reprocessed version of the image set which starts from
a camera RAW image file that has been minimally processed by
the camera and creates a linearly processed, lossless 12-bit PNG file
using the well-known dcraw program. Each image contains a
color checker for determining ground truth.

\emph{NUS and NUS-Laboratory 8-camera image sets.}
The NUS 8-camera image set consists of 1736 images captured by
eight different cameras \cite{ChengPB2014}, where
in most cases each camera
has photographed the same scene. Each image is captured as
a minimally processed camera RAW image file and is linearly
processed to create a lossless 16-bit PNG file. Each image
contains a color checker for determining ground truth.
The NUS-Laboratory 8-camera image set is
a complementary laboratory set of 840 images captured with
the same cameras. These additional images correct
for a bias towards images captured outdoors in daylight
\cite{ChengPCB2015-ICCV}.

\begin{table}[t]
\caption{
    Comparison of accuracy of our proposed ensembles
    of multivariate regression trees against Cheng et
    al.~\cite{ChengPCB2015-CVPR} on the SFU Laboratory
    \cite{BarnardMCF2002} image set, for various performance
    measures.
}\label{TABLE:ResultsBarnard}
\centering
\begin{tabular}{@{}l | l | r r r r r}
\multicolumn{4}{l}{} &
    \multicolumn{1}{r}{tri-} &
    \multicolumn{1}{r}{best} &
    \multicolumn{1}{r}{worst} \\
meas. & meth. & mean & med. & mean & 25\% & 25\% \\\hline
%%%%%%%%%%%%%%%%%%%%%%%%%%%%%%%%%%%%%%%%%%%%%%%%%%%%%%%%%%%%%%%%%%%%%%%%%%%
\multicolumn{7}{l}{Sony DXC-930, 321 images}\\\hline
\multirow{2}{*}{$\mathit{dist}_{\mathit{\theta}}$}
&  Cheng   &    4.24 &    2.24 &    2.85 &    0.35 &    11.29 \\
&  Ours    &\hl 4.04 &\hl 1.85 &\hl 2.50 &\hl 0.25 &\hl 11.20 \\\hline
\multirow{2}{*}{$\mathit{dist}_{\mathit{rep}}$}
&  Cheng   &    4.62 &    2.68 &    3.24 &    0.43 &\yc 11.95 \\
&  Ours    &\hl 4.45 &\hl 2.40 &\hl 2.89 &\hl 0.30 &    12.03 \\\hline
%\multirow{2}{*}{$\mathit{dist}_{1} \times 10^{2}$}
%&  Cheng   &    7.48 &    4.01 &    5.03 &    0.63 &\yc 19.89 \\
%&  Ours    &\hl 7.22 &\hl 3.67 &\hl 4.53 &\hl 0.44 &    20.04 \\\hline
%\multirow{2}{*}{$\mathit{dist}_{2} \times 10^{2}$}
%&  Cheng   &    4.76 &    2.57 &    3.21 &    0.40 &    12.63 \\
%&  Ours    &\hl 4.58 &\hl 2.35 &\hl 2.91 &\hl 0.28 &\hl 12.60 \\\hline
\multirow{2}{*}{$\mathit{dist}_{\mathit{ped}}$}
&  Cheng   &    2.07 &    1.18 &    1.45 &    0.19 &\yc  5.35 \\
&  Ours    &\hl 1.99 &\hl 1.04 &\hl 1.33 &\hl 0.14 &     5.38
%%%%%%%%%%%%%%%%%%%%%%%%%%%%%%%%%%%%%%%%%%%%%%%%%%%%%%%%%%%%%%%%%%%%%%%%%%%
\end{tabular}
\end{table}

\subsection{Experimental methodology}

The main goal of our experiments is to compare our
multivariate regression tree method---where each tree
simultaneously predicts all three chromaticity components of
an illuminant and the trees are fit by directly minimizing
a distance measure of interest---to Cheng
et al.'s~\cite{ChengPCB2015-CVPR} univariate tree method.
To that end, we followed Cheng et al.'s~\cite{ChengPCB2015-CVPR}
original experimental setup closely. In particular, we used the
following methodology in our experimental evaluation.

% - Training and test sets:
%     - Cheng et al.'s features
%     - image sets, per camera
%     - processing images (normalization, masking out color checker, ...)
\emph{Training data generation.}
Cheng et al.~\cite{ChengPCB2015-CVPR} use four pairs of simple
features in their univariate trees (see the description in
Section \ref{SUBSECTION:Univariate}) and in our main set of
experiments, we used the same features.
In both approaches,
training and
testing is done on each camera separately; i.e., a predictive model is
built for a particular model of camera rather than a generic
model that can be used by any camera.
To construct the machine learning data for a camera, each image
taken by the camera in an image set is processed by
(i) normalizing the image to a [0, 1] image,
using the saturation level
and darkness level of the camera,
(ii) masking out the
saturated pixels and the color checker, if present,
(iii) determining the feature values for the image, and
(iv) labeling the example using the ground truth illuminant.
The result is
a set of labeled training examples
$\{(\vec{x}_1, \vec{y}_1), \ldots, (\vec{x}_n, \vec{y}_n)\}$, where
$\vec{x}_i = (f_{r}^{1}, f_{g}^{1}, \ldots,
f_{r}^{4}, f_{g}^{4})$
is a vector of feature values
and $\vec{y}_i = (r_{i}, g_{i}, b_{i})$ is the ground
truth illuminant.

\begin{table}[t]
\caption{
    Comparison of accuracy of our proposed ensembles
    of multivariate regression trees against Cheng
    et al.~\cite{ChengPCB2015-CVPR} on the Gehler-Shi
    \cite{ShiF,GehlerRBMS2008} image set, for various
    distance measures.
}\label{TABLE:ResultsGehlerShi}
\centering
\begin{tabular}{@{}l | l | r r r r r}
\multicolumn{4}{l}{} &
    \multicolumn{1}{r}{tri-} &
    \multicolumn{1}{r}{best} &
    \multicolumn{1}{r}{worst} \\
meas. & meth. & mean & med. & mean & 25\% & 25\% \\\hline
%%%%%%%%%%%%%%%%%%%%%%%%%%%%%%%%%%%%%%%%%%%%%%%%%%%%%%%%%%%%%%%%%%%%%%%%%%%
\multicolumn{7}{l}{Canon 1D, 86 images}\\\hline
\multirow{2}{*}{$\mathit{dist}_{\mathit{\theta}}$}
&  Cheng   &    3.74 &    2.83 &    2.91 &\yc 0.74 &    8.30 \\
&  Ours    &\hl 3.57 &\hl 2.45 &\hl 2.75 &    0.81 &\hl 8.13 \\\hline
\multirow{2}{*}{$\mathit{dist}_{\mathit{rep}}$}
&  Cheng   &    4.62 &    3.59 &    3.73 &\yc 0.89 &   10.10 \\
&  Ours    &\hl 4.38 &\hl 3.21 &\hl 3.53 &    0.92 &\hl 9.66 \\\hline
%\multirow{2}{*}{$\mathit{dist}_{1} \times 10^{2}$}
%&  Cheng   &    5.97 &    4.66 &    4.78 &\yc 1.21 &    13.03 \\
%&  Ours    &\hl 5.71 &\hl 4.02 &\hl 4.45 &    1.31 &\hl 12.88 \\\hline
%\multirow{2}{*}{$\mathit{dist}_{2} \times 10^{2}$}
%&  Cheng   &    3.94 &    3.03 &    3.11 &\yc 0.78 &    8.70 \\
%&  Ours    &\hl 3.75 &\hl 2.57 &\hl 2.91 &    0.85 &\hl 8.49 \\\hline
\multirow{2}{*}{$\mathit{dist}_{\mathit{ped}}$}
&  Cheng   &    1.58 &    1.23 &    1.27 &\yc 0.36 &\yc 3.40 \\
&  Ours    &\hl 1.53 &\hl 1.10 &\hl 1.20 &    0.37 &    3.41 \\\hline
\multicolumn{7}{l}{Canon 5D, 482 images}\\\hline
\multirow{2}{*}{$\mathit{dist}_{\mathit{\theta}}$}
&  Cheng   &    2.18 &    1.37 &    1.54 &    0.35 &    5.42 \\
&  Ours    &\hl 2.06 &\hl 1.20 &\hl 1.38 &\hl 0.27 &\hl 5.32 \\\hline
\multirow{2}{*}{$\mathit{dist}_{\mathit{rep}}$}
&  Cheng   &    2.85 &    1.77 &    2.00 &    0.42 &    7.13 \\
&  Ours    &\hl 2.68 &\hl 1.54 &\hl 1.77 &\hl 0.34 &\hl 6.96 \\\hline
%\multirow{2}{*}{$\mathit{dist}_{1} \times 10^{2}$}
%&  Cheng   &    3.57 &    2.30 &    2.56 &    0.58 &    8.74 \\
%&  Ours    &\hl 3.37 &\hl 1.99 &\hl 2.27 &\hl 0.45 &\hl 8.59 \\\hline
%\multirow{2}{*}{$\mathit{dist}_{2} \times 10^{2}$}
%&  Cheng   &    2.33 &    1.47 &    1.64 &    0.37 &    5.77 \\
%&  Ours    &\hl 2.19 &\hl 1.27 &\hl 1.46 &\hl 0.29 &\hl 5.63 \\\hline
\multirow{2}{*}{$\mathit{dist}_{\mathit{ped}}$}
&  Cheng   &    0.96 &    0.63 &    0.70 &    0.17 &    2.30 \\
&  Ours    &\hl 0.92 &\hl 0.57 &\hl 0.63 &\hl 0.15 &\hl 2.29
%%%%%%%%%%%%%%%%%%%%%%%%%%%%%%%%%%%%%%%%%%%%%%%%%%%%%%%%%%%%%%%%%%%%%%%%%%%
\end{tabular}
\end{table}

% - Parameter selection:
%     - 1/3 of Gehler-Shi image set using recovery angular error
%     - stopping criteria plus other parameters
%     - 240 trees in Cheng et al. ensembles
%     - ours: 30 trees (more trees did not appear to help)
\emph{Parameter selection.}
In both approaches, parameters were set on 1/3 of the
Gehler-Shi image set using $\mathit{dist}_{\mathit{\theta}}$
as the distance measure. The parameters were then fixed for
all other cameras, image sets, and distance measures. Cheng et
al.~\cite{ChengPCB2015-CVPR} set the following parameters:
(i) the number of trees in an ensemble
($30 \times 4 \times 2 = 240$ trees, where
each feature pair is used to build a tree
for predicting $r$ and a tree for predicting $g$,
and this is repeated 30 times),
(ii) the amount of overlap and the number of slices of the
training data used in constructing the ensemble, and
(iii) a threshold value for determining the consensus of the
ensemble.
In our method, we set the following parameters:
(i) the number of trees in an ensemble (30 trees),
(ii) the amount of randomization used in constructing the ensemble
(10\%; see \ref{SUBSECTION:Ensembles}), and
(iii) a threshold value, where a node is partitioned
only if the average error at the node is greater than the threshold
(0.5).

As well, Cheng et al.~\cite{ChengPCB2015-CVPR} use the
MATLAB routine \texttt{fitrtree} for fitting a tree using the
squared-error loss function. The routine has two parameters
that were used at their default values: each branch node in the
tree has at least ``MinParentSize = 10'' examples and each leaf
has at least ``MinLeafSize = 1'' examples per tree leaf. These
values are used as a stopping criteria when building the tree.
We used the same stopping criteria in our implementation for
fitting a tree using a chosen white balancing distance measure.

\begin{table}[h!]
\caption{
    \emph{Part I:} Comparison of accuracy of our proposed ensembles
    of multivariate regression trees against Cheng
    et al.~\cite{ChengPCB2015-CVPR} on the combined
    NUS \cite{ChengPB2014}
    and
    NUS-Laboratory 8-camera \cite{ChengPCB2015-ICCV}
    image sets, for various distance measures.
}\label{TABLE:ResultsChengetalPartI}
\centering
\begin{tabular}{@{}l | l | r r r r r}
\multicolumn{4}{l}{} &
    \multicolumn{1}{r}{tri-} &
    \multicolumn{1}{r}{best} &
    \multicolumn{1}{r}{worst} \\
meas. & meth. & mean & med. & mean & 25\% & 25\% \\\hline
%%%%%%%%%%%%%%%%%%%%%%%%%%%%%%%%%%%%%%%%%%%%%%%%%%%%%%%%%%%%%%%%%%%%%%%%%%%
\multicolumn{7}{l}{Canon EOS-1Ds Mark III, 364 images}\\\hline
\multirow{2}{*}{$\mathit{dist}_{\mathit{\theta}}$}
&   Cheng   &    2.13 &    1.45 &    1.61 &    0.36 &\yc 5.06 \\
&   Ours    &\hl 2.03 &\hl 1.25 &\hl 1.46 &\hl 0.26 &    5.09 \\\hline
\multirow{2}{*}{$\mathit{dist}_{\mathit{rep}}$}
&   Cheng   &    2.81 &    1.89 &    2.14 &    0.46 &    6.71 \\
&   Ours    &\hl 2.64 &\hl 1.56 &\hl 1.88 &\hl 0.35 &\hl 6.60 \\\hline
%\multirow{2}{*}{$\mathit{dist}_{1} \times 10^{2}$}
%&   Cheng   &    3.56 &    2.41 &    2.68 &    0.61 &\yc 8.42 \\
%&   Ours    &\hl 3.41 &\hl 2.11 &\hl 2.46 &\hl 0.45 &    8.50 \\\hline
%\multirow{2}{*}{$\mathit{dist}_{2} \times 10^{2}$}
%&   Cheng   &    2.29 &    1.56 &    1.73 &    0.39 &\yc 5.42 \\
%&   Ours    &\hl 2.19 &\hl 1.34 &\hl 1.57 &\hl 0.29 &    5.49 \\\hline
\multirow{2}{*}{$\mathit{dist}_{\mathit{ped}}$}
&  Cheng    &    1.00 &    0.70 &    0.76 &    0.19 &\yc 2.32 \\
&  Ours     &\hl 0.96 &\hl 0.60 &\hl 0.69 &\hl 0.14 &    2.37 \\\hline
%%%%%%%%%%%%%%%%%%%%%%%%%%%%%%%%%%%%%%%%%%%%%%%%%%%%%%%%%%%%%%%%%%%%%%%%%%%
\multicolumn{7}{l}{Canon EOS 600D, 305 images}\\\hline
\multirow{2}{*}{$\mathit{dist}_{\mathit{\theta}}$}
&   Cheng   &\yc 2.21 &\yc 1.55 &\yc 1.67 &    0.45 &\yc 5.12 \\
&   Ours    &    2.26 &    1.61 &    1.74 &\hl 0.33 &    5.30 \\\hline
\multirow{2}{*}{$\mathit{dist}_{\mathit{rep}}$}
&   Cheng   &\yc 2.92 &\yc 2.04 &\yc 2.23 &    0.57 &\yc 6.76 \\
&   Ours    &    2.96 &    2.07 &    2.29 &\hl 0.43 &    6.92 \\\hline
%\multirow{2}{*}{$\mathit{dist}_{1} \times 10^{2}$}
%&   Cheng   &\yc 3.65 &\yc 2.55 &\yc 2.79 &    0.76 &\yc 8.44 \\
%&   Ours    &    3.76 &    2.72 &    2.93 &\hl 0.56 &    8.71 \\\hline
%\multirow{2}{*}{$\mathit{dist}_{2} \times 10^{2}$}
%&   Cheng   &\yc 2.37 &\yc 1.67 &\yc 1.80 &    0.49 &\yc 5.48 \\
%&   Ours    &    2.42 &    1.74 &    1.87 &\hl 0.36 &    5.63 \\\hline
\multirow{2}{*}{$\mathit{dist}_{\mathit{ped}}$}
&  Cheng    &\yc 1.01 &    0.72 &    0.78 &    0.23 &\yc 2.32 \\
&  Ours     &    1.02 &    0.72 &    0.78 &\hl 0.15 &    2.40 \\\hline
%%%%%%%%%%%%%%%%%%%%%%%%%%%%%%%%%%%%%%%%%%%%%%%%%%%%%%%%%%%%%%%%%%%%%%%%%%%
\multicolumn{7}{l}{Fujifilm X-M1, 301 images}\\\hline
\multirow{2}{*}{$\mathit{dist}_{\mathit{\theta}}$}
&   Cheng   &\yc 2.17 &    1.35 &    1.54 &    0.31 &\yc 5.39 \\
&   Ours    &    2.19 &\hl 1.21 &\hl 1.46 &\hl 0.23 &    5.72 \\\hline
\multirow{2}{*}{$\mathit{dist}_{\mathit{rep}}$}
&   Cheng   &\yc 2.94 &    1.77 &    2.07 &    0.40 &\yc 7.37 \\
&   Ours    &    3.00 &\hl 1.60 &\hl 1.99 &\hl 0.32 &    7.94 \\\hline
%\multirow{2}{*}{$\mathit{dist}_{1} \times 10^{2}$}
%&   Cheng   &\yc 3.57 &    2.22 &    2.55 &    0.52 &\yc 8.85 \\
%&   Ours    &    3.65 &\hl 2.03 &\hl 2.47 &\hl 0.40 &    9.56 \\\hline
%\multirow{2}{*}{$\mathit{dist}_{2} \times 10^{2}$}
%&   Cheng   &\yc 2.34 &    1.45 &    1.66 &    0.34 &\yc 5.81 \\
%&   Ours    &    2.36 &\hl 1.30 &\hl 1.59 &\hl 0.25 &    6.18 \\\hline
\multirow{2}{*}{$\mathit{dist}_{\mathit{ped}}$}
&  Cheng    &    1.04 &    0.66 &    0.75 &    0.17 &\yc 2.54 \\
&  Ours     &    1.04 &\hl 0.59 &\hl 0.72 &\hl 0.12 &    2.72 \\\hline
%%%%%%%%%%%%%%%%%%%%%%%%%%%%%%%%%%%%%%%%%%%%%%%%%%%%%%%%%%%%%%%%%%%%%%%%%%%
\multicolumn{7}{l}{Nikon D5200, 305 images}\\\hline
\multirow{2}{*}{$\mathit{dist}_{\mathit{\theta}}$}
&   Cheng   &     2.07 &    1.30 &    1.47 &    0.39 &\yc 5.02 \\
&   Ours    &\hl  2.03 &\hl 1.15 &\hl 1.34 &\hl 0.30 &    5.18 \\\hline
\multirow{2}{*}{$\mathit{dist}_{\mathit{rep}}$}
&   Cheng   &     2.92 &    1.89 &    2.11 &    0.49 &\yc 7.11 \\
&   Ours    &\hl  2.87 &\hl 1.64 &\hl 1.95 &\hl 0.40 &    7.36 \\\hline
%\multirow{2}{*}{$\mathit{dist}_{1} \times 10^{2}$}
%&   Cheng   &     3.51 &    2.21 &    2.52 &    0.67 &\yc 8.47 \\
%&   Ours    &\hl  3.47 &\hl 2.01 &\hl 2.33 &\hl 0.52 &    8.80 \\\hline
%\multirow{2}{*}{$\mathit{dist}_{2} \times 10^{2}$}
%&   Cheng   &     2.27 &    1.44 &    1.62 &    0.43 &\yc 5.48 \\
%&   Ours    &\hl  2.22 &\hl 1.29 &\hl 1.49 &\hl 0.33 &    5.64 \\\hline
\multirow{2}{*}{$\mathit{dist}_{\mathit{ped}}$}
&  Cheng    &     1.03 &    0.66 &    0.76 &    0.23 &\yc 2.43 \\
&  Ours     &\hl  0.99 &\hl 0.59 &\hl 0.69 &\hl 0.16 &    2.46
%%%%%%%%%%%%%%%%%%%%%%%%%%%%%%%%%%%%%%%%%%%%%%%%%%%%%%%%%%%%%%%%%%%%%%%%%%%
\end{tabular}
\end{table}

\begin{table}[h!]
\caption{
    \emph{Part II:} Comparison of accuracy of our proposed ensembles
    of multivariate regression trees against Cheng
    et al.~\cite{ChengPCB2015-CVPR} on the combined
    NUS \cite{ChengPB2014}
    and
    NUS-Laboratory 8-camera \cite{ChengPCB2015-ICCV}
    image sets, for various distance measures.
}\label{TABLE:ResultsChengetalPartII}
\centering
\begin{tabular}{@{}l | l | r r r r r}
\multicolumn{4}{l}{} &
    \multicolumn{1}{r}{tri-} &
    \multicolumn{1}{r}{best} &
    \multicolumn{1}{r}{worst} \\
meas. & meth. & mean & med. & mean & 25\% & 25\% \\\hline
%%%%%%%%%%%%%%%%%%%%%%%%%%%%%%%%%%%%%%%%%%%%%%%%%%%%%%%%%%%%%%%%%%%%%%%%%%%
\multicolumn{7}{l}{Olympus E-PL6, 313 images}\\\hline
\multirow{2}{*}{$\mathit{dist}_{\mathit{\theta}}$}
&   Cheng   &    1.77 &    1.11 &    1.25 &    0.26 &\yc 4.38 \\
&   Ours    &    1.77 &\hl 1.02 &\hl 1.18 &\hl 0.23 &    4.56 \\\hline
\multirow{2}{*}{$\mathit{dist}_{\mathit{rep}}$}
&   Cheng   &\yc 2.45 &    1.49 &    1.70 &    0.36 &\yc 6.15 \\
&   Ours    &    2.46 &\hl 1.39 &\hl 1.60 &\hl 0.32 &    6.41 \\\hline
%\multirow{2}{*}{$\mathit{dist}_{1} \times 10^{2}$}
%&   Cheng   &\yc 2.96 &    1.86 &    2.10 &    0.45 &\yc 7.29 \\
%&   Ours    &    2.99 &\hl 1.73 &\hl 2.00 &\hl 0.41 &    7.68 \\\hline
%\multirow{2}{*}{$\mathit{dist}_{2} \times 10^{2}$}
%&   Cheng   &\yc 1.92 &    1.20 &    1.36 &    0.29 &\yc 4.75 \\
%&   Ours    &    1.93 &\hl 1.10 &\hl 1.27 &\hl 0.26 &    4.96 \\\hline
\multirow{2}{*}{$\mathit{dist}_{\mathit{ped}}$}
&  Cheng   &    0.87 &    0.56 &    0.63 &    0.15 &\yc 2.11 \\
&  Ours    &\hl 0.86 &\hl 0.50 &\hl 0.58 &\hl 0.13 &    2.20 \\\hline
%%%%%%%%%%%%%%%%%%%%%%%%%%%%%%%%%%%%%%%%%%%%%%%%%%%%%%%%%%%%%%%%%%%%%%%%%%%
\multicolumn{7}{l}{Panasonic Lumix DMC-GX1, 308 images}\\\hline
\multirow{2}{*}{$\mathit{dist}_{\mathit{\theta}}$}
&   Cheng   &    2.22 &    1.41 &    1.57 &    0.36 &\yc 5.47 \\
&   Ours    &\hl 2.12 &\hl 1.17 &\hl 1.35 &\hl 0.23 &    5.63 \\\hline
\multirow{2}{*}{$\mathit{dist}_{\mathit{rep}}$}
&   Cheng   &    3.02 &    1.92 &    2.17 &    0.45 &\yc 7.44 \\
&   Ours    &\hl 2.88 &\hl 1.53 &\hl 1.87 &\hl 0.29 &    7.68 \\\hline
%\multirow{2}{*}{$\mathit{dist}_{1} \times 10^{2}$}
%&   Cheng   &    3.65 &    2.37 &    2.63 &    0.62 &\yc 8.89 \\
%&   Ours    &\hl 3.48 &\hl 1.99 &\hl 2.26 &\hl 0.38 &    9.11 \\\hline
%\multirow{2}{*}{$\mathit{dist}_{2} \times 10^{2}$}
%&   Cheng   &    2.39 &    1.53 &    1.70 &    0.40 &\yc 5.88 \\
%&   Ours    &\hl 2.30 &\hl 1.29 &\hl 1.48 &\hl 0.25 &    6.08 \\\hline
\multirow{2}{*}{$\mathit{dist}_{\mathit{ped}}$}
&  Cheng   &    1.03 &    0.68 &    0.75 &    0.21 &\yc 2.44 \\
&  Ours    &\hl 0.98 &\hl 0.59 &\hl 0.66 &\hl 0.13 &    2.50 \\\hline
%%%%%%%%%%%%%%%%%%%%%%%%%%%%%%%%%%%%%%%%%%%%%%%%%%%%%%%%%%%%%%%%%%%%%%%%%%%
\multicolumn{7}{l}{Samsung NX2000, 307 images}\\\hline
\multirow{2}{*}{$\mathit{dist}_{\mathit{\theta}}$}
&   Cheng   &    2.16 &    1.38 &    1.52 &    0.39 &\yc 5.33 \\
&   Ours    &\hl 2.15 &\hl 1.30 &\hl 1.47 &\hl 0.33 &    5.45 \\\hline
\multirow{2}{*}{$\mathit{dist}_{\mathit{rep}}$}
&   Cheng   &    2.96 &    1.85 &    2.07 &    0.49 &\yc 7.38 \\
&   Ours    &\hl 2.94 &\hl 1.71 &\hl 1.97 &\hl 0.42 &    7.55 \\\hline
%\multirow{2}{*}{$\mathit{dist}_{1} \times 10^{2}$}
%&   Cheng   &    3.69 &    2.39 &    2.63 &    0.68 &\yc 9.03 \\
%&   Ours    &\hl 3.68 &\hl 2.24 &\hl 2.53 &\hl 0.56 &    9.29 \\\hline
%\multirow{2}{*}{$\mathit{dist}_{2} \times 10^{2}$}
%&   Cheng   &    2.36 &    1.52 &    1.68 &    0.44 &\yc 5.79 \\
%&   Ours    &\hl 2.35 &\hl 1.43 &\hl 1.62 &\hl 0.36 &    5.93 \\\hline
\multirow{2}{*}{$\mathit{dist}_{\mathit{ped}}$}
&  Cheng   &    1.08 &    0.73 &    0.80 &    0.23 &\yc 2.55 \\
&  Ours    &    1.08 &\hl 0.71 &\hl 0.77 &\hl 0.19 &    2.62 \\\hline
%%%%%%%%%%%%%%%%%%%%%%%%%%%%%%%%%%%%%%%%%%%%%%%%%%%%%%%%%%%%%%%%%%%%%%%%%%%
\multicolumn{7}{l}{Sony SLT-A57, 373 images}\\\hline
\multirow{2}{*}{$\mathit{dist}_{\mathit{\theta}}$}
&   Cheng   &    1.99 &    1.37 &    1.49 &    0.38 &    4.68 \\
&   Ours    &\hl 1.86 &\hl 1.09 &\hl 1.29 &\hl 0.28 &\hl 4.65 \\\hline
\multirow{2}{*}{$\mathit{dist}_{\mathit{rep}}$}
&   Cheng   &    2.76 &    1.88 &    2.05 &    0.49 &    6.55 \\
&   Ours    &\hl 2.54 &\hl 1.49 &\hl 1.78 &\hl 0.38 &\hl 6.37 \\\hline
%\multirow{2}{*}{$\mathit{dist}_{1} \times 10^{2}$}
%&   Cheng   &    3.41 &    2.35 &    2.58 &    0.66 &    7.99 \\
%&   Ours    &\hl 3.18 &\hl 1.83 &\hl 2.22 &\hl 0.50 &\hl 7.94 \\\hline
%\multirow{2}{*}{$\mathit{dist}_{2} \times 10^{2}$}
%&   Cheng   &    2.18 &    1.50 &    1.64 &    0.42 &    5.12 \\
%&   Ours    &\hl 2.05 &\hl 1.20 &\hl 1.43 &\hl 0.32 &\hl 5.10 \\\hline
\multirow{2}{*}{$\mathit{dist}_{\mathit{ped}}$}
&  Cheng   &    0.98 &    0.68 &    0.75 &    0.21 &\yc 2.26 \\
&  Ours    &\hl 0.94 &\hl 0.55 &\hl 0.66 &\hl 0.16 &    2.34
%%%%%%%%%%%%%%%%%%%%%%%%%%%%%%%%%%%%%%%%%%%%%%%%%%%%%%%%%%%%%%%%%%%%%%%%%%%
\end{tabular}
\end{table}

% - Evaluation:
%     - 10-fold cross-validation
%     - 30 experiments
\emph{Performance evaluation.}
We used standard $k$-fold cross validation,
to evaluate and compare the accuracy of our method and Cheng
et al.'s~\cite{ChengPCB2015-CVPR} method.
In $k$-fold cross validation, the
image set is randomly partitioned into $k$ approximately equal
folds and each of the folds is, in turn, used as a testing set
and the remaining $k-1$ folds are used as a training set. In
our experiments we used 10-fold cross validation, as 10-fold
is the most widely recommended, especially for our setting where
the amount of data per camera is limited in some regions
of the output space \cite{Kohavi1995,BengioG2004,HastieTF2009}.
To reduce variance, the statistics we report are the result of
performing 30 runs of 10-fold cross validation with different
random seeds. Both methods
share the same random seeds so that
the partitions into training and test are the same for
each algorithm for each experiment. Although for space
considerations we only directly compare against Cheng et
al.'s~\cite{ChengPCB2015-CVPR} method, we note that results
comparing \cite{ChengPCB2015-CVPR} to
many other algorithms can be found in their
original paper (the results reported here differ somewhat
to those reported in \cite{ChengPCB2015-CVPR} on common image
sets as that paper uses 3-fold cross validation and
reports results based on only a single run).

\subsection{Experimental results}
\label{SUBSECTION:ExperimentalResults}

We compared the two approaches on accuracy, simplicity, and
speed. All experiments were performed on a PC with an Intel
i7-6700K, 4GHz running MATLAB R2016a.

\emph{Accuracy.}
Tables~\ref{TABLE:ResultsBarnard}--\ref{TABLE:ResultsChengetalPartII}
show a comparison of the accuracy of our method of
ensembles of multivariate regression trees against Cheng et
al.'s~\cite{ChengPCB2015-CVPR} method on three distance
measures (for space reasons results for distance measures
$\mathit{dist}_{1}$ and
$\mathit{dist}_{2}$
are omitted).
For uniformity of presentation, the results for
$\mathit{dist}_{\mathit{ped}}$ are multiplied by $10^{2}$.
Note that, for each camera,
in Cheng et al.'s~\cite{ChengPCB2015-CVPR} method an ensemble is
fit once using the squared error and then evaluated on each of the
distance measures, whereas in our method an ensemble is fit and
evaluated on each of the distance measures. In practice, of course,
in our method one would choose the distance measure that best fit
with the intended purpose.
It has been noted that
the mean is a poor summary statistic in this context and more
appropriate statistical measures are the median and the trimean
\cite{GijsenijGL2009,HordleyF2006}.
We also report the mean
of the best 25\% errors, and the mean of the worst 25\% errors.
Figure~\ref{FIGURE:histogram} summarizes the percentage improvement
in accuracy for the median of all the distance measures.
On these image sets, our method
has improved accuracy for 10 out of 11 cameras and 50 out of 55 combinations
of camera and distance measure. Only for one camera does our method lead to
a decrease in accuracy and the percentage decrease in the median
is bounded by 4.5\%.

\begin{figure}[thb]
\centering
\begin{tabular}{@{}cc@{}}
\placeimage[0.45]{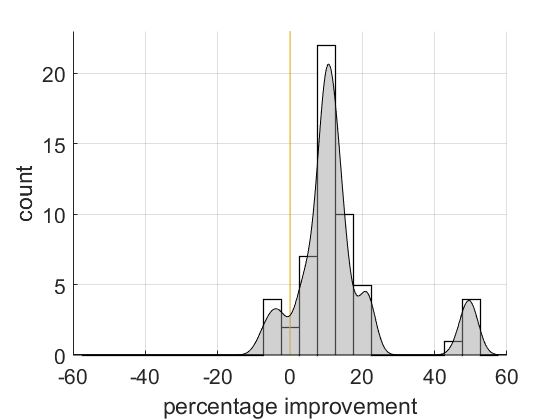} &
    \placeimage[0.45]{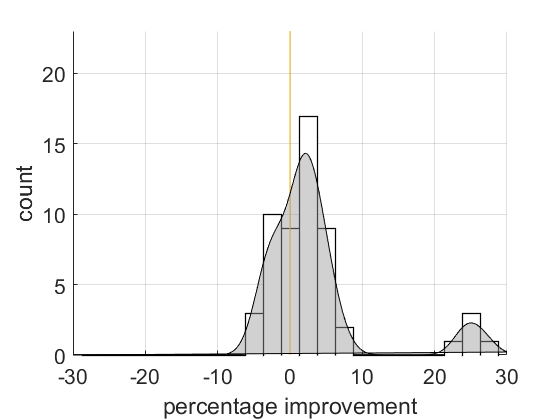} \\
(a) &
    (b)
\end{tabular}
\caption{
    %Histogram of percentage improvement of our method over Cheng
    Percentage improvement of our method over Cheng
    et al.~\cite{ChengPCB2015-CVPR} when adding the feature
    shades-of-gray for (a) the median and (b) the mean of the
    worst 25\% of the distance measures for an image set, where
    each image set is captured by a single model of a camera, and
    there are 11 image sets and 5 distance measures for each image
    set. For 50 (/55) times our method gave an improvement in the
    median and for 38 (/55) times our method gave an improvement
    in the mean of the worst 25\%.
}\label{FIGURE:SG}
\end{figure}

As can be seen in Figure~\ref{FIGURE:histogram} and
Tables~\ref{TABLE:ResultsBarnard}--\ref{TABLE:ResultsChengetalPartII},
our method compares favorably on the two statistics---median
and trimean---that are arguably the most important in white
balancing. Nevertheless, it can also be seen that Cheng et
al.'s~\cite{ChengPCB2015-CVPR} method consistently outperforms
our method on the mean of the worst 25\% errors and it has
been contended that reducing worst-case performance is also
important \cite{Mosny2012}.
In the experiments reported so far, we followed Cheng et
al.'s~\cite{ChengPCB2015-CVPR} original experimental setup
closely including using the same features. However, an
examination of why our method did more poorly on the mean of
the worst 25\% errors suggested that our method was running
out of predictive features when building our multivariate trees.
We thus pursued a secondary set of
experiments where we added one additional pair of features:
($f_{r}^{sg}$, $f_{g}^{sg}$),
the color chromaticity as
provided by the shades-of-gray algorithm
\cite{FinlaysonT2004}.
The shades-of-gray algorithm is based on the $l_{p}$-norm,
for $p = 5$.
Figure~\ref{FIGURE:SG} summarizes the results and it can be seen
that the addition of the feature improved the
performance of our method on the worst cases, albeit at the expense
of reducing the improvement on the median.
% No systematic exploration of additional features was performed,
% and we leave such an investigation to future work.

\emph{Simplicity.}
We also compared the size of the trees constructed by both
methods. For the Canon 1Ds Mark III camera, we recorded the
size of each tree and the size of each ensemble of trees, as
measured by the number of nodes. The following table
shows averages over 30 runs.
\begin{center}
\begin{tabular}{l|l|r|r}
method & measure & tree & ensemble \\\hline
Cheng & all & 206.6 & 49,575.0 \\\hline
\multirow{5}{*}{Ours}
& $\mathit{dist}_{\mathit{\theta}}$ & 143.9 &  4,317.2 \\
& $\mathit{dist}_{\mathit{rep}}$    & 153.6 &  4,609.0 \\
& $\mathit{dist}_{1}$               & 153.3 &  4,598.4 \\
& $\mathit{dist}_{2}$               & 147.8 &  4,435.0 \\
& $\mathit{dist}_{\mathit{ped}}$    & 110.9 &  3,327.6
\end{tabular}
\end{center}
It can be seen that our method leads to ensembles
that are an order of magnitude smaller.

% \emph{Simplicity.}
% We also compared the size of the trees constructed by both
% methods. For the Canon 1Ds Mark III camera, we recorded the
% size of each tree and the size of each ensemble of trees, as
% measured by the number of nodes. Table~\ref{TABLE:ResultsSize}
% shows the average tree and ensemble size, as measured over
% 30 runs. It can be seen that our method leads to ensembles
% that are an order of magnitude smaller.
% 
% \begin{table}[thb]
% \caption{
%     Comparison of simplicity (nodes) of our proposed ensembles
%     of multivariate regression trees against Cheng et
%     al.~\cite{ChengPCB2015-CVPR}.
% }\label{TABLE:ResultsSize}
% \centering
% \begin{tabular}{l|l|r|r}
% method & measure & tree & ensemble \\\hline
% Cheng & all & 206.6 & 49,575.0 \\\hline
% \multirow{5}{*}{Ours}
% & $\mathit{dist}_{\mathit{\theta}}$ & 143.9 &  4,317.2 \\
% & $\mathit{dist}_{\mathit{rep}}$    & 153.6 &  4,609.0 \\
% & $\mathit{dist}_{1}$               & 153.3 &  4,598.4 \\
% & $\mathit{dist}_{2}$               & 147.8 &  4,435.0 \\
% & $\mathit{dist}_{\mathit{ped}}$    & 110.9 &  3,327.6
% \end{tabular}
% \end{table}

\emph{Speed.}
We also compared the speed of both methods.
For the offline training of the trees,
building our multivariate trees is considerably slower.
For example, for the
Canon EOS-1Ds Mark III camera and an image set of 364 images
the times were 14.6 seconds versus 311.3 seconds.
A partial reason is that Cheng et al.~\cite{ChengPCB2015-CVPR}
use the MATLAB routine \texttt{fitrtree} for fitting a tree
which automatically parallelizes training.
For the online run-time,
the two methods are for practical purposes identical, ours
being negligibly faster due to having fewer trees in an ensemble,
but both completing in approximately 0.5 seconds or less.

We conclude the experimental evaluation with a brief comparison
to the current best-performing convolutional neural network (CNN)
approach \cite{Shi2016}.
In terms of accuracy,
on the NUS 8-camera image set,
Shi et al.~\cite{Shi2016} report
percentage improvements over
Cheng et al.~\cite{ChengPCB2015-CVPR} of
5.1\% for the mean,
8.2\% for the median, and
3.4\% for the trimean, where the improvement is
measured over the geometric means of the eight cameras.
As a point of comparison,
on the combined NUS and NUS-Laboratory 8-camera image set,
we achieve an improvement of
1.9\% for the mean,
10.7\% for the median, and
7.1\% for the trimean, again as measured over the
geometric means.
In terms of simplicity and speed,
the CNN approach is at a definite disadvantage.
For example, Shi et al.~\cite{Shi2016}
state that processing an image takes
approximately 3 seconds \emph{on a GPU}.

%
% Restate the conclusions.
% Perhaps mention future work.
%
\section{Conclusion}

We show how multivariate regression trees, where each tree
predicts multiple responses, can be used to effectively estimate
an illuminant for white balancing an image. In our proposed
method a multivariate regression tree is fit by directly
minimizing a distance measure of interest. We show empirically
that overall our method leads to improved performance on
diverse image sets. Our ensembles of multivariate regression
trees are, with some exceptions, more accurate and considerably
simpler, while inheriting the fast run-time of previous work.

%\subsection*{Acknowledgments}
%
% This work was supported in part by NSERC Discovery
% Grants. We thank x and y for their assistance on the project.
% Akshaya Senthil --- looked at algorithm selection problem.
% Weijie Wang --- looked at algorithm selection problem, found that knn
%		 gave the best performance, investigated
%		 improving performance of fmincon for exact method for
%		 find best partition when building trees.
% Michelle Pokrass --- investigate Caffe, using SharcNET for GPU computing,
%		 initial model based on RAPID paper.
% Felix Chen --- implement features f1, f2, f3 in C++/OpenCV, understand
%		 reading ground truth from the MATLAB files, add noise to
%		 an image, initial experimentation with Weka.
% Tengyu Cai --- implement feature f4 in C++/OpenCV, gray-world and
%		 white-patch algorithms, add noise to an image, initial
%		 experimentation with Weka.
%

\clearpage
\appendix
\section{Approximation error}
\label{APPENDIX:approximation}

Our proposed method is to solve
\emph{approximately} the minimization problem posed in
Equation~2 for fitting multivariate
trees (see Section 3.4).
In this appendix, we report on experiments investigating
the quality of the approximate solutions.

\begin{figure}[h!]
\centering
\begin{tabular}{cc}
\placeimage[0.45]{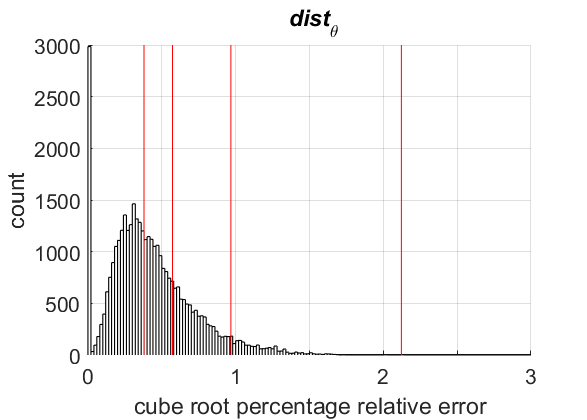} &
    \placeimage[0.45]{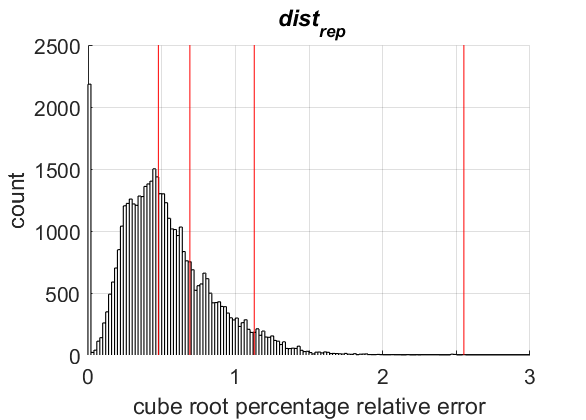} \\
\placeimage[0.45]{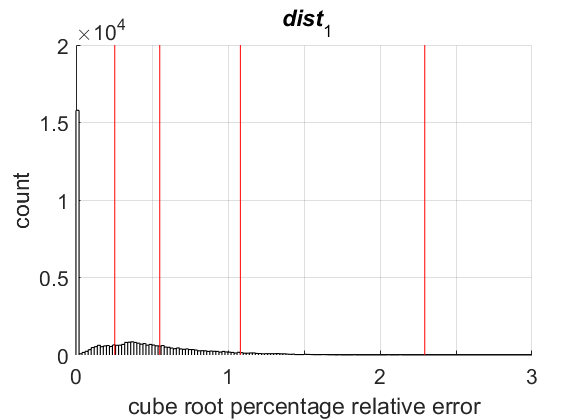} &
    \placeimage[0.45]{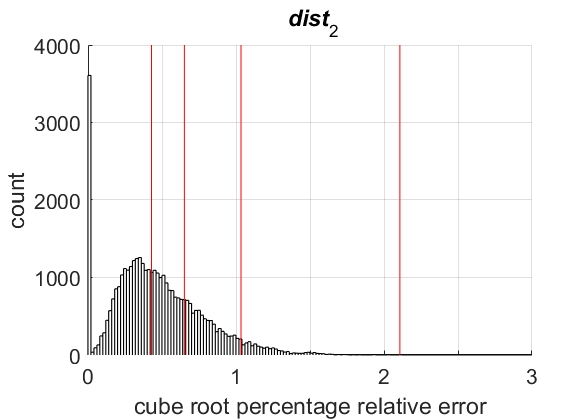} \\
\placeimage[0.45]{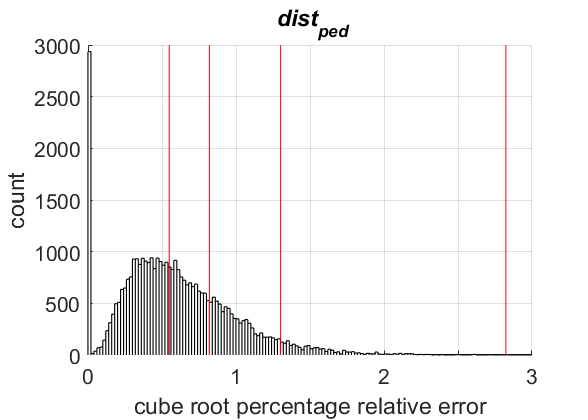}
\end{tabular}
\caption{
%Set relative error to zero if approximation = optimal = 0.
Percent relative approximation error on each call to the
minimization routine when building one tree for the Canon
EOS-1Ds Mark III camera. The red bars show the median, the
75th percentile, the 95th percentile, and the maximum. For
visualization purposes, the results were transformed by taking
the cube root of the percentage relative error.
}\label{FIGURE:approximation}
\end{figure}

For determining the exact solution to the minimization problems
we used the MATLAB routine \texttt{fmincon}.
Figure~\ref{FIGURE:approximation} shows how close to optimal
the approximate solutions were for the various distance measures.
The majority of the approximate solutions are within 1\% of optimal.

\clearpage
%\bibliography{focus}
%\bibliographystyle{plain}

\end{document}

%% file: exampleTree.tex
\tikzstyle{level 1}=[level distance=2.75cm, sibling distance=3.75cm]
\tikzstyle{level 2}=[level distance=2.75cm, sibling distance=2.00cm]
\tikzstyle{level 3}=[level distance=2.75cm, sibling distance=1.00cm]
\tikzstyle{int}  = [circle, minimum width=6pt, draw=black, inner sep=0pt]
\tikzstyle{leaf} = [circle, minimum width=6pt, fill, green, draw=black, inner sep=0pt]
\centering
\begin{tikzpicture}[grow=right, sloped, scale=0.90, every node/.style={scale=0.90}]
\node[int] {}
    child {
        node[int] {}
            child {
                node[int] {}
                    child {
                        node[int, label=right:
                            {$\cdots$}] {}
                        edge from parent
                        node[below] {\small{$\mathbf{f_{r}^{1}} > 0.4191$}}
                    }
                    child {
                        node[leaf, label=right:
                            {\small{4: (0.3622, 0.4727, 0.1651)}}] {}
                        edge from parent
                        node[above] {\small{$\mathbf{f_{r}^{1}} \le 0.4191$}}
                    }
                edge from parent
                node[below] {\small{$\mathbf{f_{r}^{4}} > 0.4078$}}
            }
            child {
                node[int] {}
                    child {
                        node[int, label=right:
                            {$\cdots$}] {}
                        edge from parent
                        node[below] {\small{$\mathbf{f_{r}^{4}} > 0.2902$}}
                    }
                    child {
                        node[leaf, label=right:
                            {\small{8: (0.2897, 0.4674, 0.2429)}}] {}
                        edge from parent
                        node[above] {\small{$\mathbf{f_{r}^{4}} \le 0.2902$}}
                    }
                edge from parent
                node[above] {\small{$\mathbf{f_{r}^{4}} \le 0.4078$}}
            }
            edge from parent
            node[below] {\small{$\mathbf{f_{r}^{1}} > 0.3297$}}
    }
    child {
        node[int] {}
            child { node[int] {}
                    child {
                        node[int, label=right:
                            {$\cdots$}] {}
                        edge from parent
                        node[below] {\small{$\mathbf{f_{r}^{3}} > 0.2840$}}
                    }
                    child {
                        node[leaf, label=right:
                            {\small{8: (0.2503, 0.4779, 0.2718)}}] {}
                        edge from parent
                        node[above] {\small{$\mathbf{f_{r}^{3}} \le 0.2840$}}
                    }
                edge from parent
                node[below] {\small{$\mathbf{f_{r}^{2}} > 0.2658$}}
            }
            child { node[int] {}
                    child {
                        node[int, label=right:
                            {$\cdots$}] {}
                        edge from parent
                        node[below] {\small{$\mathbf{f_{g}^{4}} > 0.4549$}}
                    }
                    child {
                        node[int, label=right:
                            {$\cdots$}] {}
                        edge from parent
                        node[above] {\small{$\mathbf{f_{g}^{4}} \le 0.4549$}}
                    }
                edge from parent
                node[above] {\small{$\mathbf{f_{r}^{2}} \le 0.2658$}}
            }
            edge from parent
            node[above] {\small{$\mathbf{f_{r}^{1}} \le 0.3297$}}
    };
\end{tikzpicture}